\documentclass[pmlr]{jmlr}

\usepackage{longtable}

\usepackage{hyperref}
\hypersetup{
    colorlinks=true,
    linkcolor=blue,
    filecolor=magenta,      
    urlcolor=cyan,
}
\usepackage{booktabs}
\usepackage{float}

\usepackage[load-configurations=version-1]{siunitx} 

\theorembodyfont{\upshape}
\theoremheaderfont{\scshape}
\theorempostheader{:}
\theoremsep{\newline}

\title[DeepSPINE]{DEEP SPINE: AUTOMATED LUMBAR VERTEBRAL SEGMENTATION, DISC-LEVEL DESIGNATION, AND SPINAL STENOSIS GRADING USING DEEP LEARNING}

\author{\Name{Jen-Tang Lu$^1$} \Email{jlu24@partners.org}
		\AND
        \Name{Stefano Pedemonte$^1$} \Email{stefano.pedemonte@gmail.com}
        \AND
        \Name{Bernardo Bizzo$^1$} \Email{bbizzo@mgh.harvard.edu}
        \AND
        \Name{Sean Doyle$^1$} \Email{sdoyle@mgh.harvard.edu}
        \AND
        \Name{Katherine P. Andriole$^{1,2}$} \Email{kandriole@bwh.harvard.edu}
        \AND
        \Name{Mark H. Michalski$^1$} \Email{mmichalski1@partners.org}
        \AND
        \Name{R. Gilberto Gonzalez$^3$} 			 \Email{rggonzalez@mgh.harvard.edu}
        \AND
        \Name{Stuart R. Pomerantz$^{1,3}$}  \Email{spomerantz@mgh.harvard.edu} \\
\\
       \addr $^1$MGH $\&$ BWH Center for Clinical Data Science, Boston, MA, USA\\
       \addr $^2$Department of Radiology, Brigham and Women's Hospital, Boston, MA, USA\\
       \addr $^3$Division of Neuroradiology, Massachusetts General Hospital, Boston, MA, USA} 

\begin{document}

\maketitle

\begin{abstract}
  The high prevalence of spinal stenosis results in a large volume of MRI imaging, yet interpretation can be time-consuming with high inter-reader variability even among the most specialized radiologists. In this paper, we develop an efficient methodology to leverage the subject-matter-expertise stored in large-scale archival reporting and image data for a deep-learning approach to fully-automated lumbar spinal stenosis grading. Specifically, we introduce three major contributions: (1) a natural-language-processing scheme to extract level-by-level ground-truth labels from free-text radiology reports for the various types and grades of spinal stenosis  (2) accurate vertebral segmentation and disc-level localization using a U-Net architecture combined with a spine-curve fitting method, and (3) a multi-input, multi-task, and multi-class convolutional neural network to perform central canal and foraminal stenosis grading on both axial and sagittal imaging series inputs with the extracted report-derived labels applied to corresponding imaging level segments. This study uses a large dataset of 22796 disc-levels extracted from 4075 patients. We achieve state-of-the-art performance on lumbar spinal stenosis classification and expect the technique will increase both radiology workflow efficiency and the perceived value of radiology reports for referring clinicians and patients.
\end{abstract}

\section{Introduction}

Degenerative lumbar spinal stenosis is a major cause of low back pain and is one of the most common indications for spinal surgery \citep{deyo2010trends}. The high prevalence of degenerative spinal diseases, especially in the working-age population, leads to large societal costs from related treatment and disability \citep{fayssoux2010indirect}. A significant portion of these costs comes from medical imaging which is utilized for both initial diagnosis and follow-up evaluation in both conservative and surgical treatment pathways. Magnetic resonance imaging (MRI) is the imaging modality of choice to evaluate spinal stenosis due to its superior ability to characterize soft tissue detail such as neural tissue. It allows radiologists to identify the location, etiology, and severity of nerve root compression and generate a report with level-by-level detail communicating these findings to referring physicians and their patients. Imaging reports thus inform clinical decision-making with regard to different therapeutic approaches and also are used to assess treatment response. Interpretation of spine MRI can be very time-consuming especially when advanced multi-level degeneration is present, which is not uncommon in the elderly population. Unfortunately, a lack of universally accepted imaging-based grading system or diagnostic criteria for spinal stenosis leads to large inter-reader variability even among specialists which can degrade the perceived  value of their reporting  \citep{fu2014interrater}.

To address the challenges of spinal imaging interpretation, a variety of computer-aided diagnosis techniques have been explored over the past decade for potential applicability.   Many of the previous works used computer vision techniques, such as histogram of oriented gradients \citep{ghosh2012new,oktay2013simultaneous, lootus2014vertebrae}, probabilistic models \citep{corso2008lumbar,aslan2010automated,raja2011labeling}, and GrowCut \citep{egger2017vertebral}, to perform segmentation or localize vertebral bodies and discs. For automated diagnosis of spinal degeneration, \citet{koompairojn2010computer} used hand-crafted features of T2-axial images and multilayer perceptron for stenosis classification. \citet{zhang2017weakly} proposed a weakly-supervised approach to extract salient features for lumbar spinal stenosis in MRI using window filters and pathological labels only. Spurred by recent advances in graphics-processing-unit (GPU) technology, machine learning techniques have received much attention of late.  Specifically, an appreciation of the applicability of convolutional neural networks (CNN) has led to so-called \textit{deep learning} approaches where algorithms automatically learn representative features from raw data at multiple different levels of abstraction to perform classification tasks at a high level of performance \citep{gulshan2016development, bejnordi2017diagnostic,esteva2017dermatologist}. Recently, the U-Net \citep{ronneberger2015u} deep learning algorithm has proven to be effective in segmentation tasks even with limited data and has been used for vertebral segmentation in spinal CT \citep{janssens2018fully} and X-ray imaging\citep{al2017shape}. Another good example is the work of \citet{jamaludin2017spinenet, jamaludin2017issls} in which a multi-task VGG-M architecture was developed for Pfirrmann grading of disc degeneration, central canal stenosis, and other spinal diseases. 

\begin{figure}[H]
  \centering 
  \includegraphics[width=4.8in]{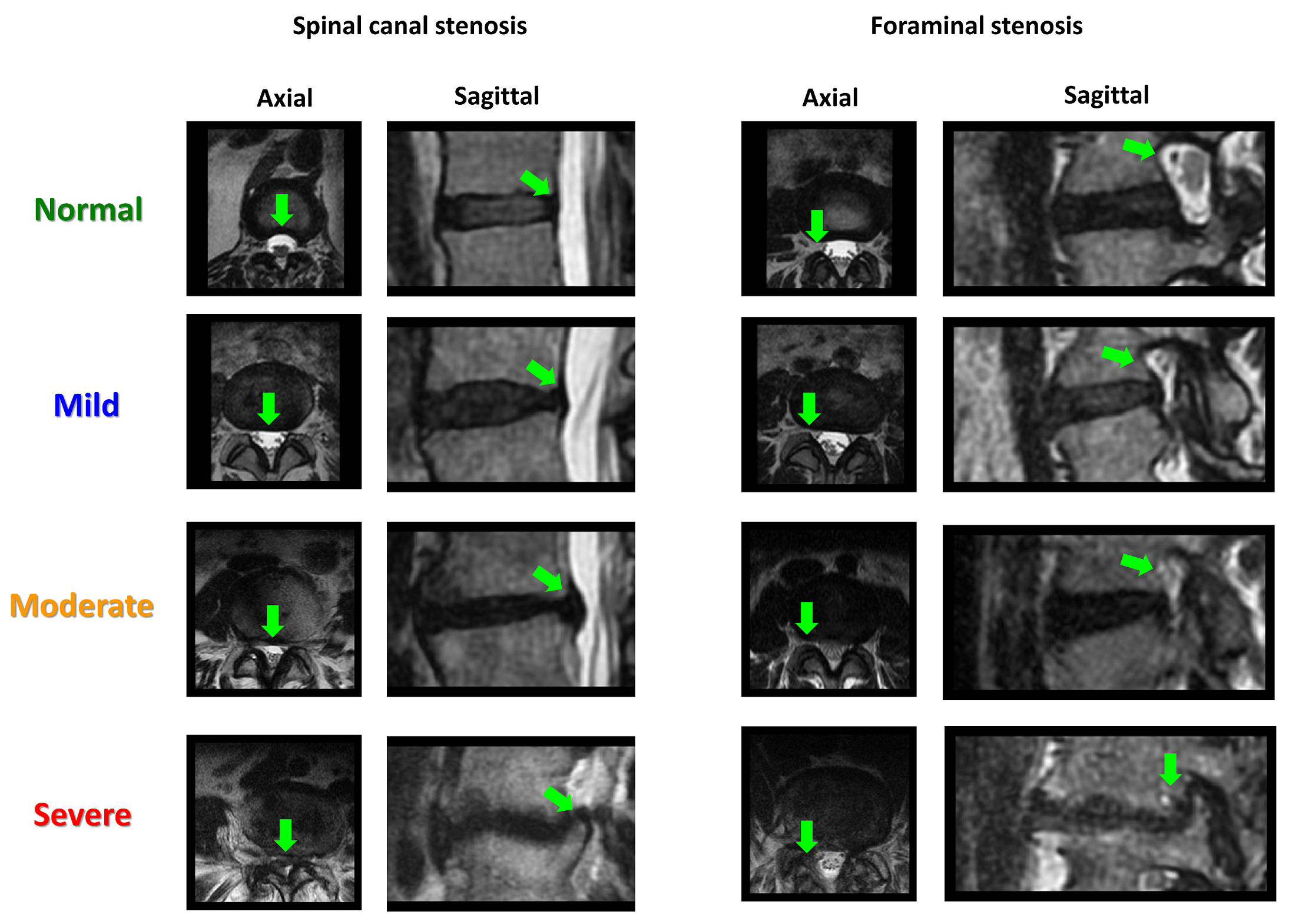} 
  \caption{Examples of different severities of stenosis at the central canal and foramen.}
  \label{fig:severity_examples} 
\end{figure}

Some published works for diagnosing spinal stenosis have had only binary classification as their goal i.e. disease-absent (negative) versus disease-present (positive) only. Yet, the radiologist's reporting task requires a description of the severity of stenosis to guide clinical decision-making. Neuroimaging and musculoskeletal radiologists at our institution predominantly use a 4-point system (normal, mild, moderate, and severe) to grade spinal stenosis severity for the central canal and bilateral foramina at each lumbar spinal level. Previous work on computer-aided diagnosis of spinal stenosis used only axially- or only sagittally-oriented MRI slices as imaging inputs \citep {zhang2017weakly, jamaludin2017spinenet,  jamaludin2017issls}. However, the interpretation of spinal MRI typically includes simultaneous review of both orientations as different anatomical planes often provide complementary information, particularly with the non-isotropic scan slice resolution typically acquired in clinical practice. Figure~\ref{fig:severity_examples} shows examples of varying grades of spinal canal stenosis and foraminal stenosis severity on representative slices from sagittal and axial imaging series.

\paragraph{Technical Significance} In this paper, we aim to utilize deep neural networks trained on large-scale archival reporting and image data for automated spine-level labeling and stenosis grading. Specifically, we use a U-Net architecture combined with a spine-curve fitting technique for vertebral body and disc-level segmentation and labeling. We then develop a \textit{multi-input} (sagittal and axial imaging), \textit{multi-task} (central canal, right foraminal, and left foraminal stenosis), and \textit{multi-class} (normal, mild, moderate, and severe) CNN for stenosis grading. A highly-supervised approach to training typically requires a large volume of manually labeled image data \citep{gulshan2016development}. However, acquiring this annotation from expert readers at sufficient scale can be prohibitive, especially with the complexity of spinal MRI where there may be 18 or more specific locations graded in a typical report. Instead, we use natural language processing (NLP) of the free text spinal MRI reports in our institution's radiology reporting archive to extract the necessary per-level stenosis grades for each imaging study. These can then be applied as location- and level-specific labels to the corresponding spinal imaging segmentations for subsequent training of the stenosis grading architecture. This work thus outlines an efficient and accurate method for medical institutions to leverage large scale archival data when applying machine learning methods to complex medical tasks.

\paragraph{Clinical Relevance} The high prevalence of spinal stenosis in the working-age and elderly population results in heavy utilization of spinal MRI in medical care, yet reporting can be time-consuming with high inter-reader variability. Automated stenosis grading will likely play an increasingly important role in spinal MRI interpretation by enabling radiologists to provide more accurate and consistent assessments in less time.  Additionally, the resultant data-rich layer beneath the standard free-text radiology report enables more useful visual representations of the described anatomy and pathology for referring physicians and their patients as well as provides ``big data" for population-based longitudinal analyses that can better inform allocation of healthcare resources.

\section{Cohort}
The imaging and reporting datasets were from consecutive non-contrast MRI examinations of the lumbar spine performed between April 2016 and October 2017 by Massachusetts General Hospital (MGH) Department of Radiology at its  inpatient, emergency, and outpatient imaging facilities.  The non-contrast imaging protocol is typically used for evaluation of low back pain and/or radiculopathy and thus excluded the majority of patients undergoing evaluation for potential spinal trauma, infection, inflammation, tumors, or post-surgical evaluation for which gadolinium-contrast is typically administered. 

\paragraph{Data Characteristics} Our initial cohort consisted of 7108 lumbar spine MRI examinations reviewed by 57 different final-signing radiologists. For each study, the sagittal and axial T2-weighted MR series were utilized for segmentation and subsequent algorithm training. The images were acquired using MR scanners of at least 1.5T  strength (GE Healthcare and Siemens Medical Imaging) and included imaging slices from both conventional fast-spin-echo acquisitions with slice thickness in the 3 - 4.0 mm range and high-resolution volumetric technique with slice thickness down to 0.9 mm. For algorithm training, we included studies for which the report text parsing was complete for explicit descriptions of all 6 spinal levels (T12-L1 through L5-S1) and the imaging segmentation was successful for at least the lower 5 levels (L1-2 through L5-S1). These criteria resulted in 22,796 intervertebral disc-levels in 4075 patients from the original cohort extracted for neural network training and testing.

\paragraph{Parsing of radiology report text for weakly-supervised learning} Many computer-aided diagnosis techniques have been developed as fully-supervised models trained on manual pixel-level delineation of regions of abnormality in the images. Such strongly-labeled data is prohibitively hard to obtain at large scale due to the practical constraints of acquiring such intensive markup from expert readers. A less labor-intensive approach is where class labels, such as presence or absence of disease or its severity, are applied to imaging in a weakly-supervised fashion. In our implementation of this approach,  labels denoting severity of spinal stenosis were extracted using NLP of free-text medical reports in which such findings were localized by spinal level enabling application to corresponding disc-level image segmentation. 

\begin{figure}[H]
  \centering 
  \includegraphics[width=4.8in]{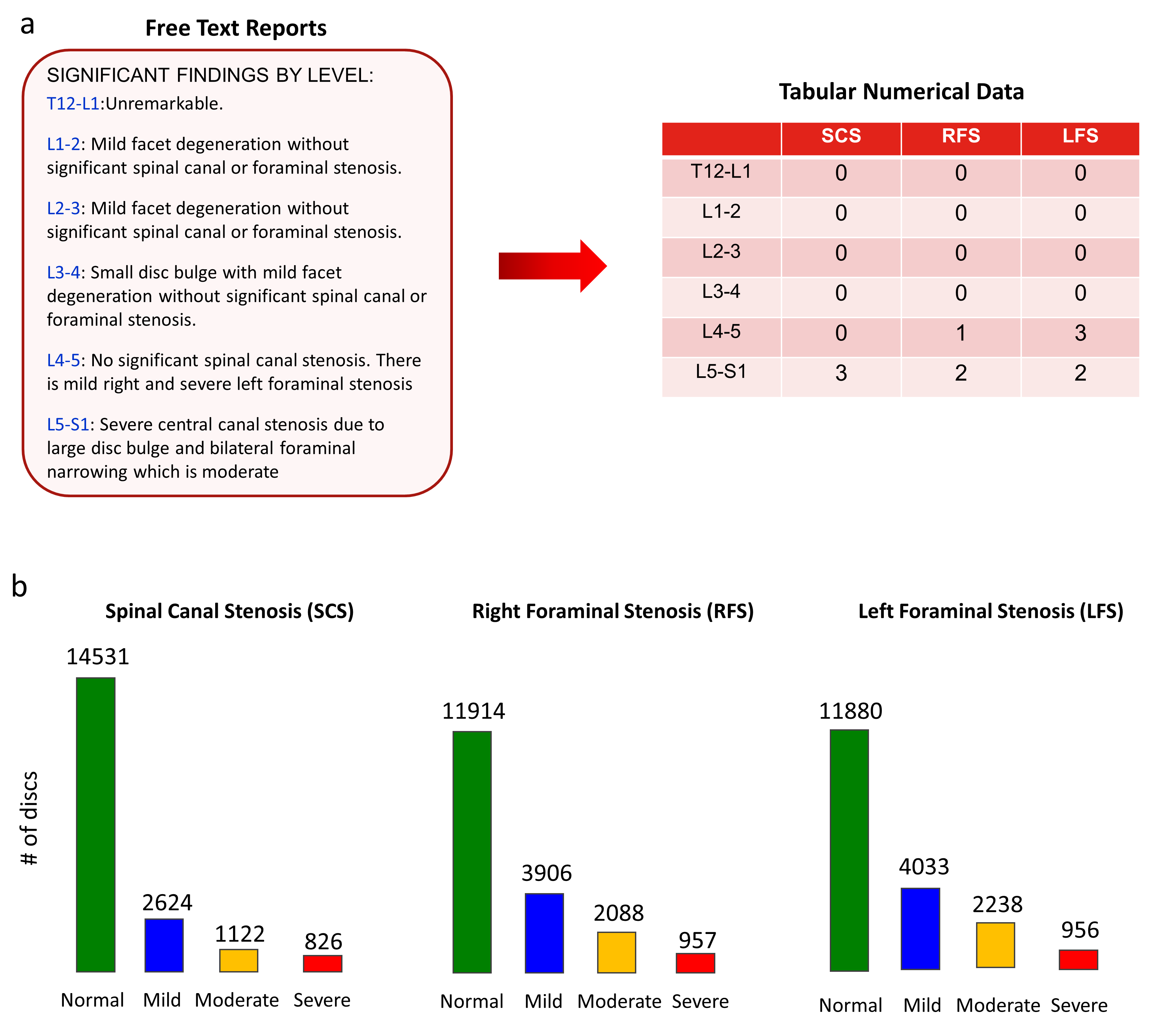} 
  \caption{Radiology report parsing and stenosis grading distribution. (a) A natural-language processing pipeline to convert the highly variable free text in the reports into numerical tabular labels for model training (0:normal, 1:mild, 2:moderate, 3:severe). (b) Stenosis grading distribution of the dataset (22796 disc-levels in total; some of them might contain only one or two labels).}
  \label{fig:data_summary} 
\end{figure}

At each disc-level of the lumbar spine, the central spinal canal and both right and left neural foramina (where the spinal nerves exit the canal) are characterized by the radiologist for any degree of degenerative stenosis. Typical textual descriptors utilized for the absence of clinically-significant disease include ``normal" or ``unremarkable."  Similarly, common descriptors for characterizing increasing severity or grade of stenosis are utilized such as  ``mild", ``moderate", ``severe" and, to a lesser extent, intermediate grades of ``mild-moderate" and ``moderate-severe". To obtain discrete stenosis grade labels for model training, we utilized regular-expression matching to extract free text descriptors and consistently map them to ordinal numerical values (0: normal/no-significant-disease, 1:mild, 2:moderate, 3:severe). At each spinal level, such stenosis grade values were extracted for spinal canal stenosis (SCS), right foraminal stenosis (RFS), and left foraminal stenosis (LFS) (Figure~\ref{fig:data_summary}a). Due to their relative lower utilization, the intermediate grades were grouped with the higher grade i.e. mild-moderate descriptors as \textit{2:moderate} and moderate-severe as \textit{3:severe}.
As is often the case for medical data, the stenosis dataset is highly imbalanced with a preponderance of disease in the normal and mild categories with relatively less  higher grade disease. The detailed distribution of stenosis grading in the dataset is shown in Figure~\ref{fig:data_summary}b.

A challenging aspect of the regular-expression matching was the large number of synonyms utilized by the large group of radiology report authors for the same spinal anatomy and pathologic terms. Another challenge was the highly variable and intermixed ordering of the description of the three types of stenosis at each spinal level in different free-text sentence constructions which often also included the etiology of stenosis such as disc bulging and hypertrophic facet degeneration. Examples are shown as below:
\newline
\newline
\noindent
\textbf{Synonymous term variability:}
\begin{itemize}
\item Normal
\begin{itemize}
\item Normal
\item Unremarkable
\item Without significant spinal canal or foraminal stenosis
\end{itemize}
\item Stenosis
\begin{itemize}
\item Stenosis
\item Narrowing
\item Compromise
\item Triangulation
\item Nerve root encroachment
\item Neural impingement
\end{itemize}
\item Central Canal (stenosis)
\begin{itemize}
\item Central canal
\item Central spinal canal
\item Central spinal
\item Spinal canal
\item Central zone
\item Central 
\item Canal
\end{itemize}
\item Neural Foramen (stenosis)
\begin{itemize}
\item Neural foramen
\item Neuro-foramen
\item Neuroforamen
\item Foramen
\item Neuroforaminal 
\end{itemize}
\end{itemize}

\noindent 
\textbf{Complex and highly variable sentence construction. Examples:}
\begin{itemize}
\item There is no significant central canal stenosis and mild right and moderate left foraminal narrowing.
\item Moderate right and mild left stenosis are present. No evidence of spinal canal narrowing is observed.
\item Severe canal stenosis and bilateral foraminal narrowing which is severe as well. 
\end{itemize}

\section{Methods}

\subsection{Segmentation and labeling of intervertebral disc-levels}
\begin{figure}[H]
  \centering 
  \includegraphics[width=6.0in]{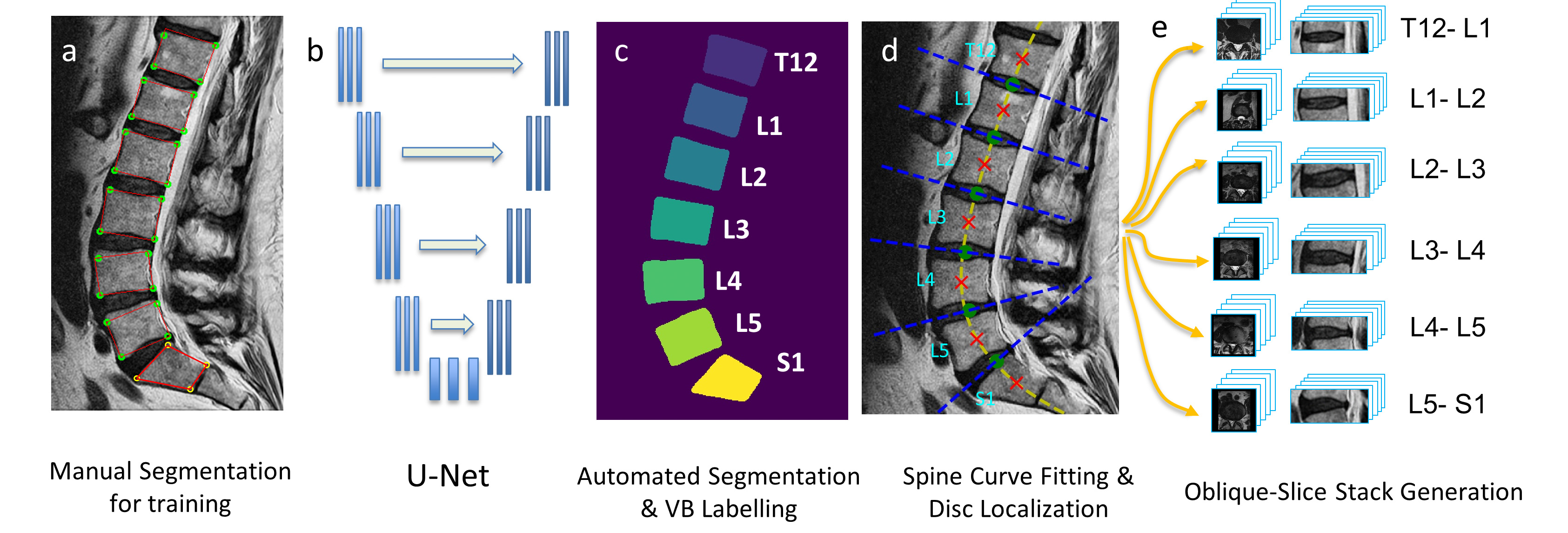} 
  \caption{Workflow of vertebral segmentation: vertebral body labeling and disc-level image volume extraction. (a) Ground truth creation: manually generating bounding boxes for lumbar and sacral vertebral bodies. (b) U-Net ($512 \times 512$ input/output with 5 downsampling/upsampling modules, 2 convolutional layers per module, and 32 initial features in the network) for vertebral segmentation. (c) Predicted vertebral segmentation and spine-level label assignment. (d) Spine curve fitting (yellow line), disc (green dots) and vertebrae (red crosses) localization, and disc plane localization (blue lines). (e) Disc-level image volume extraction in both sagittal and axial view.}
  \label{fig:unet_workflow} 
\end{figure}

To be able to apply extracted stenosis grade labels to correct anatomic locations in the imaging data, segmentation of the 6 intervertebral disc-levels (T12-L1 through L5-S1) was necessary. Instead of directly localizing discs, we started from vertebral body segmentation since vertebral body contours are more consistent than the disc appearance especially with superimposed degenerative disc disease. 
The network for vertebral body segmentation was based on a U-Net architecture. We added batch normalization before each ReLU activation and used the sigmoid function for the final activation to generate a pixel-wise probability map for the segmentation. We performed segmentation on the central slices of the sagittal T2-weighted series and created ground-truth masks of vertebral body contour by manually marking the four corners of the vertebral body on an individual slice and generating the corresponding bounding boxes (Fig. \ref{fig:unet_workflow}a). 1000 cases were randomly selected from the cohort with a 60:20:20 split to train, validate, and test the model. As lumbar and sacral vertebra have different shapes, two detectors were trained independently, one for the vertebra of T12-L5 (relatively rectangular) and the other one for the sacral S1 segment (relatively trapezoidal). Both the lumbar and sacral detectors were trained with stochastic gradient descent (SGD) with a learning rate of 0.01 and negative Dice coefficient \citep{milletari2016v} as the loss function:
\begin{equation}
L_{seg} = - \frac{2\sum_{i}^{N}p_ig_i+ \epsilon}{\sum_{i}^{N}p_i + \sum_{i}^{N}g_i + \epsilon}
\end{equation}

\noindent
where the sums run over the N pixels of the predicted segmentation $p_i\in$P and the ground-truth binary masks of $g_i\in$G; $\epsilon$ is a small number (set to 1.0 in the experiment). The predicted lumbar and sacral segmentations were then combined, and the corresponding labels were assigned to each vertebral level (Fig. \ref{fig:unet_workflow}).

To extract image volumes aligned to individual disc planes, the spinal curvature was first approximated with a polynomial curve fit to the centroids of the predicted vertebral bodies. The locations of intervertebral discs were then assigned to the midpoints between the centroids of the two consecutive vertebra, and the disc planes were approximated to the lines perpendicular to the tangent lines of the spine curve at the disc points. For more accurate assessment of the spinal canal contour, we then 
extracted 3D image volumes in both sagittal and axial planes oriented to the exact plane of each intervertebral disc. The axial views were resampled with a dimension of $360 \times 360 \times 8$, where 8 represents the number of slices (corresponding to a disc volume of $9 \times 9 \times 1.6$ cm$^3$). Sagittal views were resampled with a dimension of $160 \times 320 \times 25$ (corresponding to a volume of $4 \times 8 \times 5$ cm$^3$). Each disc-level image volume was then normalized with mean subtraction.

\subsection{Algorithm training of spinal canal and foraminal stenosis grading}
\begin{figure}[htbp]
  \centering 
  \includegraphics[width=5.5in]{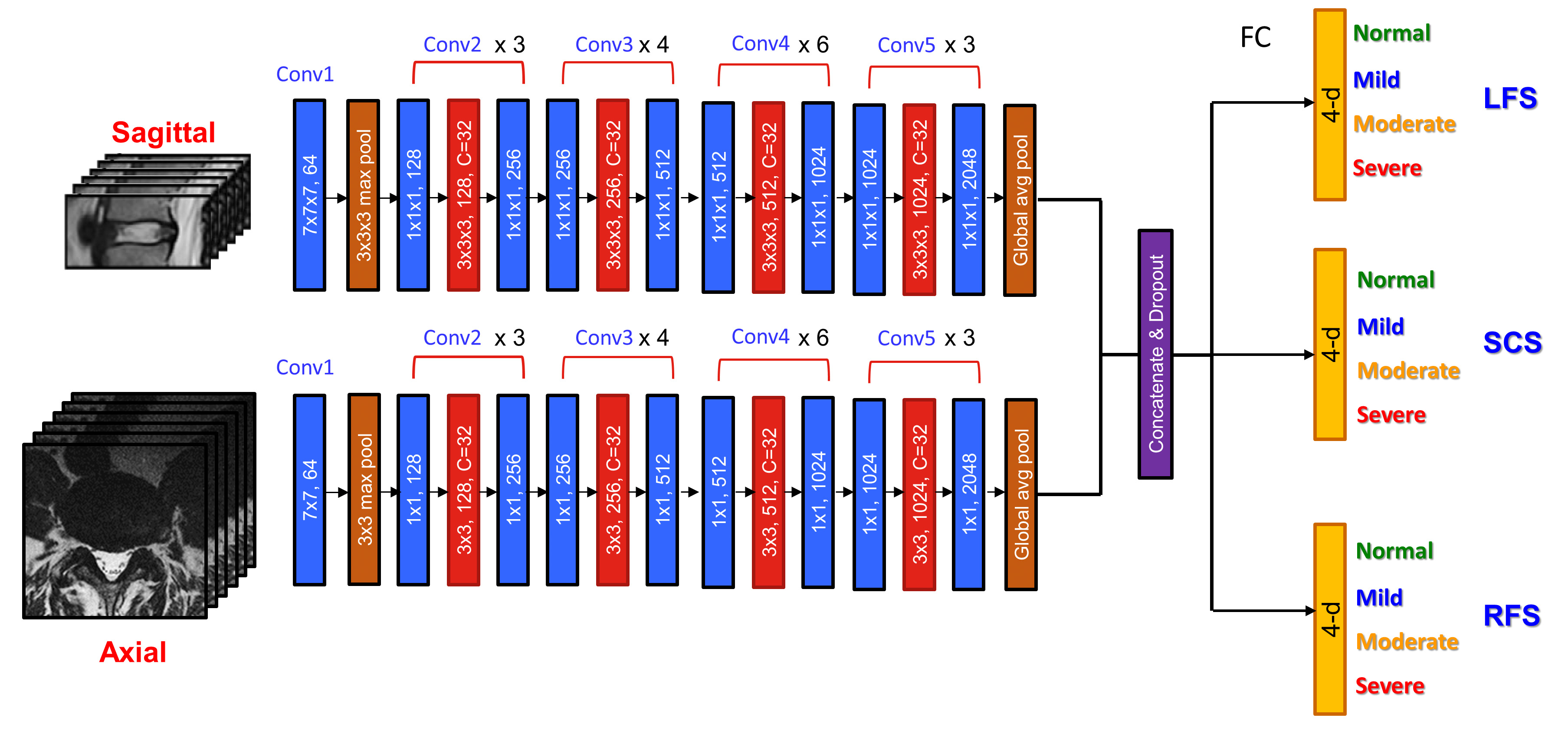} 
  \caption{Multi-input, multi-task, and multi-class version of ResNeXt-50 for stenosis grading. Inputs are re-oriented sagittal and axial disc image volumes. 3D convolutions are used for the sagittal branch and 2D convolutions are used for the axial branch. Numbers in the blue convolutional blocks represent filter size and number. Numbers in the red convolutional blocks are filter size, total number of filters, and cardinality. Conv2 to Conv5 are residual blocks. All the convolutional layers are followed by a batch normalization layer and a leaky ReLU activation layer. The two input branches are concatenated after the global average pooling layers and then split out for three different tasks: spinal canal stenosis (SCS), right foraminal stenosis (RFS), and left foraminal stenosis (LFS). Each task is performed by a fully-connected (FC) layer with the softmax activation over the four outputs to generate the probability for each stenosis grading.}
  \label{fig:network} 
\end{figure}

After the disc image volumes are extracted, they are passed to a second network for stenosis grading. We randomly split all the discs mentioned in Section 2 into training (15957 discs, 70\%), validation (3419, 15\%), and test (3420, 15\%) sets. The architecture for stenosis grading was a \textit{multi-input}, \textit{multi-task}, and \textit{multi-class} neural network. As shown in Figure ~\ref{fig:network}, there were two inputs: sagittal and axial volumes for a given disc. The network was based on ResNeXt-50 with cardinality of 32 \citep{xie2017aggregated} for both inputs. We used 2D CNN for the axial input and 3D CNN for the sagittal input because there were more slices in the sagittal disc volumes. Note that filters in 2D CNN were still three-dimensional, the last dimension of which was equal to the number of channels (slices) of the input. For example, conv1 for the axial branch was a 2D convolutional layer with $7 \times 7 \times 8$ filters as there were 8 slices in the axial inputs, while 3D convolutions were applied to the sagittal branch with $7 \times 7 \times 7$ filters. The two branches were concatenated after the global pooling layer and split out again for the three classification tasks: central (spinal) stenosis and right and left foraminal stenosis grading, each of which was a fully connected layer with four output classes (normal, mild, moderate, and severe) using the standard softmax activation to predict the probability of each stenosis grading. We used dropout regularization with a dropout rate of 0.2 on the fully connected layers. As the dataset was imbalanced, we trained the network by minimizing a weighted categorical cross entropy loss: 

\begin{equation}
L = - \sum_{n=1}^{N}\sum_{t=1}^{T}\sum_{j=1}^{C}\alpha_{j,t}y_{j,t}(x_n)log(P_{j,t}(x_n))
\end{equation}

Where ${x_n}$ represents the $n^{th}$ input disc image volume; $P_{j,t}$ is the $j^{th}$ component of the output probability for the task t; $y$ is the ground truth labels with one-hot encoding; $\alpha_{j,t}$ is the weighting factor, which is inversely proportional to the class frequency in the training set for each task t; C (equal to 4) is the total number of output classes for each task; T (equal to 3) is the total number of tasks; N is the total number of disc volumes in the training data. We used the Adadelta optimizer \citep{zeiler2012adadelta} with standard parameters (lr=1.0, rho=0.95) to train the model from scratch. We then picked the model with the lowest validation error and evaluated its performance on the test set. All the models were trained utilizing the Keras deep learning library with the Tensorflow backend on NVIDIA DGX1 with V100 GPUs.

\section{Results}

\subsection{Vertebral body segmentation}
\begin{figure}[htp]
  \centering 
  \includegraphics[width=4.0in]{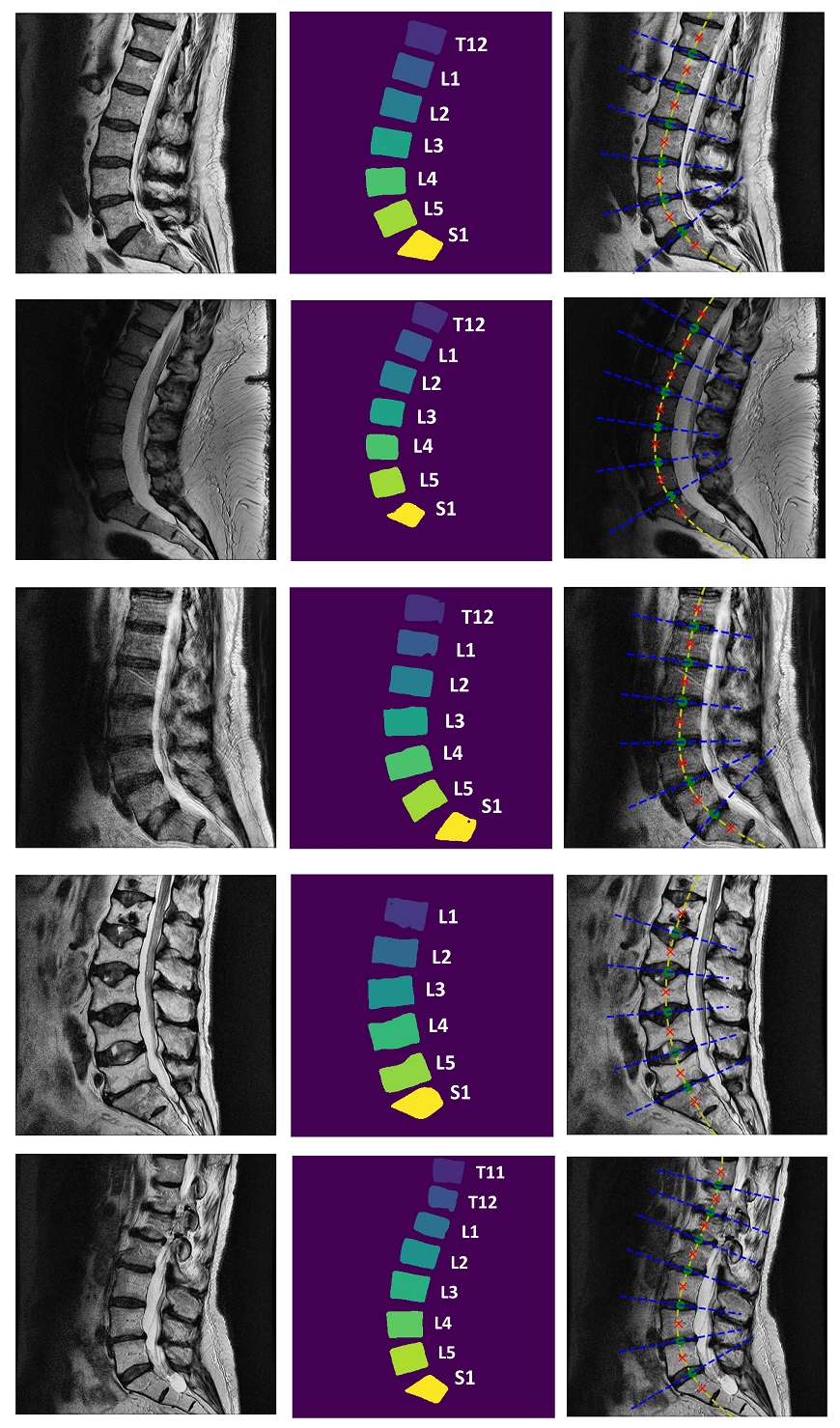} 
  \caption{Example results of vertebral body segmentation and disc localization. The left column is the input sagittal images. The central column shows the output segmentation of U-Net. The right column indicates spine curves, vertebra/disc locations, and disc planes.}
  \label{fig:seg} 
\end{figure}

We have three criteria for the success of vertebral body segmentation: (1) the detected vertebral area contains a solitary ground truth centroid of a vertebra; (2) the number of detected vertebrae is the same as that of ground truth; (3) the detected S1 segment does not overlap with the lumbar vertebra. In short, the criteria ensure that all lumbar intervertebral discs can be extracted with the algorithm. The successful rate on the test set according to these criteria is 94\% (188 out of 200 cases). The algorithm is robust for the majority of diagnostic-quality image acquisitions with exceptions including severe scoliosis and fused vertebral bodies (Fig.~\ref{fig:seg}). Other causes for segmentation failure include the most severe motion or metal-hardware artifacts, which can compromise human reader evaluation.

Besides overall success rate, we use the Dice similarity coefficient (DSC) and the centroid distance between prediction and ground truth to quantitatively characterize the performance of the algorithm. For lumbar vertebral detection, the mean DSC is 0.93 with standard deviation of 0.02, and the mean error distance between the ground truth center and the center of the detected region is 0.79 mm with standard deviation of 0.44 mm. We find a similar performance for the S1 detector with mean DSC of 0.93 (s.t.d. 0.03) and mean error distance of 0.72 mm (s.t.d. 0.47 mm). It is interesting to note that the vertebral ground-truth masks are created with bounding boxes (rectangles for T12-L5 and trapezoids for S1), and our vertebral detectors thus learn to predict bounding areas of vertebral bodies rather than tight contours. This is generally not a problem as the bounding areas allow us to accurately extract the centroids of the vertebra, which enables spine curve fitting and the extraction of disc image volumes for stenosis grading, as described in Section 3.1.

\subsection{Stenosis Grading}
For model inference and testing, stenosis grading is assigned to the output severity class with the highest probability. Table \ref{tab:table1} shows class accuracy for each stenosis grade. To get a standard deviation over the results, we trained two different models by swapping the validation and test set. When compared to ground truth labels extracted from the original radiology reports, the models had higher accuracy in classifying normal and severe stenosis than mild and moderate grades. A plausible explanation would be that there is a higher inter-reader variability for the intermediate grades than the more obvious normal and severe cases. When we combine the mild and moderate cases into one class by summing over the two output probabilities ($P_{Mild/Moderate}$= $P_{Mild}$ + $P_{Moderate}$), there is a significant improvement in performance, leading to class accuracies of greater than 80\%.

\begin{table}[H]
\centering
\caption{Class accuracy for stenosis grading.}
\label{tab:table1}
\begin{tabular}{|p{3cm}|p{3cm}|p{3cm}|p{3cm}|p{3cm}|}
\hline
\multicolumn{5}{|l|}{Spinal Canal Stenosis (\%, mean $\pm$ std)} \\ \hline
\multicolumn{1}{|c|}{Normal} & 
\multicolumn{1}{|c|}{Mild} & 
\multicolumn{1}{|c|}{Mod.} & 
\multicolumn{1}{|c|}{Severe} & 
\multicolumn{1}{|c|}{Class Avg.} 
\\ \hline
\multicolumn{1}{|c|}{\begin{tabular}[c]{@{}c@{}}78.7 $\pm$ 5.4\end{tabular}} & 
\multicolumn{1}{|c|}{\begin{tabular}[c]{@{}c@{}}59.6 $\pm$ 2.0\end{tabular}} & 
\multicolumn{1}{|c|}{\begin{tabular}[c]{@{}c@{}}61.3 $\pm$ 2.1\end{tabular}} & 
\multicolumn{1}{|c|}{\begin{tabular}[c]{@{}c@{}}82.7 $\pm$ 5.6\end{tabular}} & 
\multicolumn{1}{|c|}{\begin{tabular}[c]{@{}c@{}}70.6 $\pm$ 2.1\end{tabular}} 
\\ \hline

\multicolumn{1}{|c|}{Normal} & 
\multicolumn{2}{|c|}{Mild / Moderate} & 
\multicolumn{1}{|c|}{Severe} & 
\multicolumn{1}{|c|}{Class Avg.} 
\\ \hline
\multicolumn{1}{|c|}{\textbf{\begin{tabular}[c]{@{}c@{}}79.7 $\pm$ 3.3\end{tabular}}} & \multicolumn{2}{c|}{\textbf{\begin{tabular}[c]{@{}c@{}}83.7 $\pm$ 3.4\end{tabular}}} & \multicolumn{1}{|c|}{\textbf{\begin{tabular}[c]{@{}c@{}}77.7 $\pm$ 1.5\end{tabular}}} & \multicolumn{1}{|c|}{\textbf{\begin{tabular}[c]{@{}c@{}}80.4 $\pm$ 1.6\end{tabular}}} \\ \hline
\multicolumn{5}{|l|}{Foraminal Stenosis (\%, mean $\pm$ std)} \\ \hline

\multicolumn{1}{|c|}{Normal} & 
\multicolumn{1}{|c|}{Mild} & 
\multicolumn{1}{|c|}{Mod.} & 
\multicolumn{1}{|c|}{Severe} & 
\multicolumn{1}{|c|}{Class Avg.} \\ \hline
\multicolumn{1}{|c|}{\begin{tabular}[c]{@{}l@{}}80.5 $\pm$ 0.3\end{tabular}} & \multicolumn{1}{|c|}{\begin{tabular}[c]{@{}l@{}}61.3 $\pm$ 5.8\end{tabular}} & \multicolumn{1}{|c|}{\begin{tabular}[c]{@{}l@{}}52.0 $\pm$ 6.0\end{tabular}} & \multicolumn{1}{|c|}{\begin{tabular}[c]{@{}l@{}}74.8 $\pm$ 3.1\end{tabular}} & \multicolumn{1}{|c|}{\begin{tabular}[c]{@{}l@{}}67.1 $\pm$ 2.2\end{tabular}} \\ \hline

\multicolumn{1}{|c|}{Normal} & 
\multicolumn{2}{|c|}{Mild / Moderate} & 
\multicolumn{1}{|c|}{Severe} & 
\multicolumn{1}{|c|}{Class Avg.} \\ \hline
\multicolumn{1}{|c|}{\textbf{\begin{tabular}[c]{@{}c@{}}79.6 $\pm$ 0.8\end{tabular}}} &
\multicolumn{2}{|c|}{\textbf{\begin{tabular}[c]{@{}c@{}}84.2 $\pm$ 0.7\end{tabular}}} & \multicolumn{1}{|c|}{\textbf{\begin{tabular}[c]{@{}c@{}}70.5 $\pm$ 0.8\end{tabular}}} & \multicolumn{1}{|c|}{\textbf{\begin{tabular}[c]{@{}c@{}}78.1 $\pm$ 0.4\end{tabular}}} \\ \hline
\end{tabular}
\end{table}

We further make a comparison between models trained with axial image input only, sagittal image input only, and combined inputs. For the models with axial or sagittal input only, we turn off one or the other branch in our model. We use class average accuracy to evaluate the performance of the model for stenosis grading as the dataset is highly imbalanced. Not surprisingly, the results shown in Table \ref{tab:table2} indicate that the model trained on both sagittal and axial scans leads to better performance with 2-4\% improvement in class average accuracy and smaller standard deviation. The result also matches the fact that radiologists’ interpretations are typically based on a composite interpretation of both or more imaging series.

\begin{table}[H]
\centering
\caption{Comparison of models trained with axial input only, sagittal input only, and both inputs in class average accuracy (\%, mean $\pm$ std).}
\label{tab:table2}
\begin{tabular}{|p{4cm}|p{4cm}|p{4cm}|p{4cm}|}
\hline
\multicolumn{1}{|c|}{} & 
\multicolumn{1}{|c|}{Axial Only} & 
\multicolumn{1}{|c|}{Sagittal Only} & 
\multicolumn{1}{|c|}{Axial + Sagittal}
\\ \hline

\multicolumn{1}{|c|}{Spinal Canal Stenosis} & 
\multicolumn{1}{|c|}{78.6 $\pm$ 2.7} & 
\multicolumn{1}{|c|}{78.6 $\pm$ 2.4} & 
\multicolumn{1}{|c|}{80.4 $\pm$ 1.6}
\\ \hline

\multicolumn{1}{|c|}{Foraminal Stenosis} & 
\multicolumn{1}{|c|}{76.6 $\pm$ 2.5} & 
\multicolumn{1}{|c|}{74.3 $\pm$ 1.7} & 
\multicolumn{1}{|c|}{78.1 $\pm$ 0.4}
\\ \hline

\end{tabular}
\end{table}

In order to make a relevant comparison to previous studies that describe algorithms that perform only binary classification (normal versus stenosis), we transform our stenosis grading system into a binary classifier by combining normal, mild, and moderate stenosis classes into the negative category ($P_{neg}= P_{Normal}+P_{Mild}+P_{Moderate}$) and treating the severe class as the positive category ($P_{pos}=P_{Severe}$) due to its higher likelihood to be clinically significant. We compare the performance at different disc levels between their reported results and our model results on our data. As shown in Table \ref{tab:table3}, our model performs better at all disc-levels for both spinal canal stenosis and foraminal stenosis. We also use the area under the receiver operating characteristic curve (AUC) to quantify the performance of our binary classifier. Our model reaches an AUC of 0.983 (95\% CI, 0.971-0.992) for spinal canal stenosis and 0.961 (95\% CI, 0.955-0.967) for foraminal stenosis.

\begin{table}[H]
\centering
\caption{Comparison of the proposed algorithm with the best published results on binary classification of spinal canal and foraminal stenosis. Performance metric is overall accuracy.}
\label{tab:table3}
\begin{tabular}{|p{4cm}|p{4cm}|p{4cm}|p{4cm}|}
\hline
\multicolumn{1}{|c|}{} & 
\multicolumn{1}{|c|}{\citet{zhang2017weakly}} & 
\multicolumn{1}{|c|}{\citet{jamaludin2017spinenet}} & 
\multicolumn{1}{|c|}{\textbf{Ours}}
\\ \hline

\multicolumn{1}{|c|}{Type of Scan} & 
\multicolumn{1}{|c|}{Axial} & 
\multicolumn{1}{|c|}{Sagittal} & 
\multicolumn{1}{|c|}{\textbf{Axial + Sagittal}}
\\ \hline

\multicolumn{4}{|l|}{Spinal canal stenosis (\%, mean $\pm$ std)}
\\ \hline

\multicolumn{1}{|c|}{L3-L4} & 
\multicolumn{1}{|c|}{87.2 $\pm$ 3.2} & 
\multicolumn{1}{|c|}{\textbf{94.7}} & 
\multicolumn{1}{|c|}{\textbf{94.5 $\pm$ 0.7}}
\\ \hline

\multicolumn{1}{|c|}{L4-L5} & 
\multicolumn{1}{|c|}{85.1 $\pm$ 3.4} & 
\multicolumn{1}{|c|}{85.9} & 
\multicolumn{1}{|c|}{\textbf{95.3 $\pm$ 0.2}}
\\ \hline

\multicolumn{1}{|c|}{L5-S1} & 
\multicolumn{1}{|c|}{87.5 $\pm$ 3.3} & 
\multicolumn{1}{|c|}{93.7} & 
\multicolumn{1}{|c|}{\textbf{99.1 $\pm$ 0.5}}
\\ \hline

\multicolumn{4}{|l|}{Foraminal stenosis (\%, mean $\pm$ std)}
\\ \hline

\multicolumn{1}{|c|}{L3-L4} & 
\multicolumn{1}{|c|}{84.3 $\pm$ 3.9} & 
\multicolumn{1}{|c|}{N/A} & 
\multicolumn{1}{|c|}{\textbf{94.0 $\pm$ 0.7}}
\\ \hline

\multicolumn{1}{|c|}{L4-L5} & 
\multicolumn{1}{|c|}{84.0 $\pm$ 4.0} & 
\multicolumn{1}{|c|}{N/A} & 
\multicolumn{1}{|c|}{\textbf{89.0 $\pm$ 1.4}}
\\ \hline

\multicolumn{1}{|c|}{L5-S1} & 
\multicolumn{1}{|c|}{87.1 $\pm$ 3.4} & 
\multicolumn{1}{|c|}{N/A} & 
\multicolumn{1}{|c|}{\textbf{91.2 $\pm$ 1.6}}
\\ \hline

\end{tabular}
\end{table}

\section{Discussion}
Leveraging the large scale reporting and imaging archive at our institution, we have been able to efficiently train a highly performant deep-learning algorithm to provide automated level-by-level stenosis grading of the central canal and foramina of the lumbar spine from MR images. There are some limitations of our approach we hope to address in future work: for the segmentation task, the middle slice of the sagittal scan series was utilized for computational efficiency as it displays the most distinguishing characteristics when there is not-too-severe scoliotic curvature.  The current method is sufficient to extract disc image volumes for stenosis grading for the majority of the cases. Problematic cases of severe scoliosis will be addressed subsequently with annotations applied across an imaging volume rather than a single slice.

Our model for stenosis grading is trained from cases assessed by a large group of radiologists. Moreover, the readers are comprised of members of both neuroradiology and musculoskeletal radiology divisions within our department. Both these individual and group factors likely contribute to inter-reader variability in stenosis grading thresholds. As each ground truth label was derived from a report examined by only a single radiologist, the performance of our expert “crowd-sourced” model is thus affected by the collective precision of the heterogeneous group. We hypothesize that the output probability of the model can be analyzed in light of intra- and inter-reader variability: for an inference where there is an output class with a dominant probability (both absolute and relative to other class probabilities), we assume that the majority of radiologists would reach a consensus on that grading; in contrast, for an inference where there is no output class with a dominant probability, it probably means that radiologists' opinions may differ. To address the issue of inter-reader variability, our next step will be to perform a bias correction for the reports prior to training based on an analysis of comparative individual grade distributions. It is expected that model performance will be enhanced as the interpretations of different radiologists are normalized for similar degrees of stenosis. 

\section{Conclusion}
We have achieved high performance for automated vertebral segmentation and spinal stenosis grading in lumbar spine MRI, utilizing convolution neural networks and leveraging the ``Big Data" contained in our reporting and imaging archive. We addressed the practical barriers of recruiting fresh consensus reads at sufficient scale for algorithm training by ‘crowd-sourcing’ the contribution of a large heterogeneous group of expert radiologists through the extraction of grading labels from their prior reporting. This approach enabled us to achieve a high level of performance across the range of stenosis grading classifications for both central spinal and foraminal locations at each spinal level. It is expected that this machine learning tool will optimize clinical workflow for radiologists as a reporting aid and provide more consistent interpretation for the referring clinicians and their patients with lumbar spinal stenosis.

\acks{Many thanks to Prof. Jayashree Kalpathy-Cramer, Dr. Adrian Dalca, Dr. Christopher Bridge, Dr. Walter F. Wiggins, and Bradley Wright for valuable discussion.}

\appendix
\section*{Appendix}

Supplementary video: Level-by-level stenosis grading by DeepSPINE

\noindent
\url{https://bit.ly/DeepSPINE}

\noindent
In the video, the original sagittal scan is shown on the left and re-orientated disc volumes in oblique axial view are shown on the right with disc-levels and stenosis grading assigned by the DeepSPINE model.

\bibliography{References.bib}

\end{document}